\documentclass[11pt]{article} 
\usepackage[utf8]{inputenc}
\usepackage[english]{babel}
\usepackage[a4paper,margin=1in]{geometry} 
\usepackage{authblk} 
\usepackage[colorlinks=true, linkcolor=black, citecolor=black, urlcolor=blue, filecolor=blue,breaklinks=true]{hyperref} 
\usepackage{setspace} 
\usepackage{lineno} 
\usepackage{csquotes} 
\usepackage{ragged2e} 
\usepackage{dsfont} 
\usepackage{amsmath}
\usepackage[capitalise, noabbrev, nameinlink]{cleveref}
\crefdefaultlabelformat{#2\textbf{#1}#3} 
\Crefname{figure}{\textbf{Figure}}{\textbf{Figures}} 
\Crefname{section}{\textbf{Section}}{\textbf{Sections}} 
\Crefname{table}{\textbf{Table}}{\textbf{Tables}} 

\DeclareMathOperator*{\argmax}{arg\,max}

\usepackage{graphicx}
\graphicspath{{./main_figs/}{./supplement/suppl_figs/}} 
\DeclareGraphicsExtensions{.pdf,.jpeg,.JPG,.png,.PNG, .eps, .tiff}
\usepackage{subcaption} 
\DeclareCaptionLabelFormat{bold}{\textbf{(#2)}} 
\newcommand{\labelphantom}[1]{
  \parbox{0pt}{\phantomsubcaption\label{#1}}%
}
\usepackage{pgffor} 
\usepackage{alphalph} 
\usepackage[labelfont=bf, textfont=bf, singlelinecheck=off, textfont=footnotesize]{caption} 
\usepackage{booktabs}
\usepackage{multirow}
\usepackage{rotating}

\newcommand{\generateFigSubpanels}[6][0]{
     \expandafter\newcommand\csname#2\endcsname{ 
      \begin{figure}[htbp]
        \foreach [count=\i] \x in #6{%
            \labelphantom{fig:#2:\AlphAlph{\i}}
        }
        \centering
        \includegraphics[width=#4\linewidth,angle=#1]{#3}
        \caption{\textbf{\normalsize #5}}\label{fig:#2}
        \footnotesize
        \justifying
        \foreach [count=\i] \x in #6{%
            \noindent\subref{fig:#2:\AlphAlph{\i}}~\x\space
        }
      \end{figure}
    }
}

\newcommand{\generateFig}[6][0]{
     \expandafter\newcommand\csname#2\endcsname{ 
      \begin{figure}[htbp]
        \centering
        \includegraphics[width=#4\linewidth,angle=#1]{#3}
        \caption{\textbf{\normalsize #5}\label{fig:#2}
        \footnotesize
        \normalfont
        #6
        }
        
      \end{figure}
    }
}

\newcommand{\generateSidewaysFigSubpanels}[6][0]{
     \expandafter\newcommand\csname#2\endcsname{ 
      \begin{sidewaysfigure}[htbp]
        \foreach [count=\i] \x in #6{%
            \labelphantom{fig:#2:\AlphAlph{\i}}
        }
        \centering
        \includegraphics[width=#4\linewidth,angle=#1]{#3}
        \caption{\textbf{\normalsize #5}}\label{fig:#2}
        \footnotesize
        \justifying
        \foreach [count=\i] \x in #6{%
            \noindent\subref{fig:#2:\AlphAlph{\i}}~\x\space
        }
      \end{sidewaysfigure}
    }
}

\newcommand{\generateSidewaysFig}[6][0]{
     \expandafter\newcommand\csname#2\endcsname{ 
      \begin{sidewaysfigure}[htbp]
        \centering
        \includegraphics[width=#4\linewidth,angle=#1]{#3}
        \caption{\textbf{\normalsize #5}\label{fig:#2}
        \footnotesize
        \normalfont
        #6
        }
        
      \end{sidewaysfigure}
    }
}

\newcommand{\generateTab}[6][]{
     \expandafter\newcommand\csname#2\endcsname{ \begin{table}[ht]
          \centering
                \resizebox{#4\textwidth}{!}{
                \input{#3}
            }
            \caption{\textbf{\normalsize #5}\label{tab:#2}
            \footnotesize
            \normalfont
            #6
            }
            
        \end{table}
    }
}

\newcommand{\generateSidewaysTab}[6][]{
     \expandafter\newcommand\csname#2\endcsname{ \begin{sidewaystable}[ht]
          \centering
                \resizebox{#4\textwidth}{!}{
                \input{#3}
            }
            \caption{\textbf{\normalsize #5}\label{tab:#2}
            \footnotesize
            \normalfont
            #6
            }
            
        \end{sidewaystable}
    }
}

\usepackage[dvipsnames]{xcolor} 
\usepackage{changepage} 
\usepackage{enumitem} 

\usepackage[backend=biber,style=authoryear-comp,sorting=nyt,natbib=true, url=false]{biblatex}

\title{Do Neural Networks Generalize from Self-Averaging Sub-classifiers in the Same Way As Adaptive Boosting?
}

\author[1]{Michael Sun}
\author[2]{Peter Chatain}
\affil[1]{Stanford University, Stanford, CA}
\affil[2]{Stanford University, Stanford, CA}
\date{March 2022} 

\newcommand{\makeAbstract}{
\begin{abstract}
\noindent In recent years, neural networks (NN’s) have made giant leaps in a wide variety of domains. NN’s are often referred to as “black box” algorithms due to how little we can explain their empirical success. Our foundational research seeks to explain why neural networks generalize. A recent advancement derived a mutual information measure for explaining the performance of deep NN’s through a sequence of increasingly complex functions. We show deep NN’s learn a series of boosted classifiers whose generalization is popularly attributed to self-averaging over an increasing number of interpolating sub-classifiers. To our knowledge, we are the first authors to establish the connection between generalization in boosted classifiers and generalization in deep NN’s. Our experimental evidence and theoretical analysis suggest NNs trained with dropout exhibit similar self-averaging behavior over interpolating sub-classifiers as cited in popular explanations for the post-interpolation generalization phenomenon in boosting. 
\end{abstract}
}
\bibliography{library} 
\begin{document}

    \setstretch{1.15} 
    \maketitle
    
    \makeAbstract
    \clearpage 
    \section{Introduction}

\subsection{Introducing and Motivating the problem} \cite{hornik} showed neural networks are universal function approximators that can achieve remarkably low bias on estimating highly complex functions on high dimensional data, producing bountiful successes in a wide range of domains and applications. This makes it all the more surprising, in the view of statistical learning theory, why they can generalize well to unseen data. There have been considerable advances in demonstrating the generalization of deep neural networks. Characterizing when NNs will generalize in terms of its capacity has informed many common heuristics and practices such as dropout in the machine learning community, and has driven the rapid adoption of them in real-world applications. However, explaining why they generalize still remains an open problem.

Of particular interest is using simpler models to characterize and decompose what a NN is learning through looking at their mutual information. Boosting is an ensemble method to make a stronger classifier out of weaker classifiers, with each classifier is explainable in their features and interpretable in their decisions. In the popular variant adaptive boosting, each weak classifier is trained in succession, and the next weak classifier trains on re-weighted data which focuses on examples the ensemble does poorly on. Adaptive boosting is explainable as a weighted majority vote algorithm (described by its authors \cite{adaboost}), where each vote comes from a weak learner that has an easily explained decision rule. 
Moreover, a familiar “self-averaging” view from the original authors on Dropout has recently gained a lot of consensus for explaining why Adaboost rarely overfit in practice \cite{krieger} \cite{wyner}. Within an ensemble of classifiers, self-averaging is the process which stabilizes the fit to regions with signal, while continuing to localize the effect of noisy points on the overall fit. Similarly, dropout is often used in practice to reduce the NN from overfitting. Dropout randomly selects a subset of the neurons to ignore while training, and it selects a different subset at every epoch. When making predictions, all of the neurons are used. We are inspired by dropout to view a NN as an ensemble of sub-classifiers, and relate each iteration of boosting to some sub-network. 

The time is ripe for us to bridge these two parallel developments on generalization amongst empirical risk minimizers. This connection can go a long way in uniting two traditionally separate fields of research – boosting and deep learning – and help ground the success of neural networks in more established territory.

\subsection{Your contributions} Previous work has conjectured that neural networks learn a series of increasingly complex functions (\cite{nakkiran}), although the sequence and notion of complexity are not defined. We will define the series of increasingly complex functions as adaptive boosted classifiers with well-characterized growth in its VC dimension in terms of the number of boosting rounds. Each complexity rung from this series corresponds to a distinct phase of training.

To justify this definition, we inductively show a similarity between the learning outcomes in a single phase between a single hidden layer neural network and boosted classifier. Namely, the NN’s mutual information with $Y$, the label distribution, can be explained away by a boosted algorithm from the next phase, and the boosted algorithm’s mutual information with $Y$ can be explained away by the NN in the next phase. In addition, we run extensive experiments demonstrating that NNs learn and retain increasingly complex hypotheses from adaptive boosting. In the first phase of training, the mutual performance between a single weak learner and the NN is perfect. Then, as the NN learns increasingly complex concepts, this simple concept is not forgotten. We show that this same pattern holds in further phases of training for boosted algorithms with more weak learners. We corroborate these claims with qualitative evidence by investigating the learning order of examples for each model and find similar preference of example difficulty within the same corresponding phase. 

We further investigate the connection between boosting and neural networks by studying a chosen ensemble of sub-networks chosen as in dropout as shown by \cite{srivastava} and \cite{baldi}. In boosting, each weak learner is encouraged to perform well on different subsets of the data which decreases the correlation between each of them. We prove that training only sub-networks within our chosen ensemble similarly decreases the correlation between any two hidden neurons in a simplified setting. We also experimentally validate the self averaging behavior of NNs trained with our modified version of dropout by viewing each sub-network created by dropout as an independent learner. When compared with neural networks that do not use dropout, the self averaging effect is more pronounced.

\subsection{Related Research} There have been many attempts to formulate and motivate the search for simpler neural networks. For a short early history of the development of such techniques and how they encourage simplicity, one is encouraged to read \cite{schmidhuber}. Among these, of particular interest to us is dropout, a technique for addressing overfitting by averaging over an exponential number of “thinned” networks \cite{srivastava}. This is elaborated by \cite{baldi}'s recursive view of the averaging properties of dropout.

However, such techniques that seek to design simpler neural networks do not yield simple explanations for why the resulting models generalize.

As another attempt, there have been many attempts to explain generalization seen in broader classes of interpolating classifiers. This is the case with random forests, adaptive boosting \cite{wyner}, and nearest-neighbor schemes \cite{belkin}. Meanwhile, understanding of generalization in interpolating neural networks only began advancing recently, with theoretical analysis in more realistic settings with e.g. noisy labels and nonlinear learning dynamics by \cite{niladri} still in its nascent stages.

We seek to bridge these two complementary fields of research by showing deep NNs learn a series of boosted classifiers, whose generalization is popularly attributed to self-averaging over an increasing number of interpolating sub-classifiers. Towards that end, our survey of what’s known will overview the literature on neural network complexity and adaptive boosting separately. Then, we will address previous attempts to explain neural networks by simpler models, identify gaps in understanding, then argue why the novel connection we establish between adaptive boosting and neural networks can fill in some of those gaps.

    \clearpage 








\generateFigSubpanels{figureSuggestions}{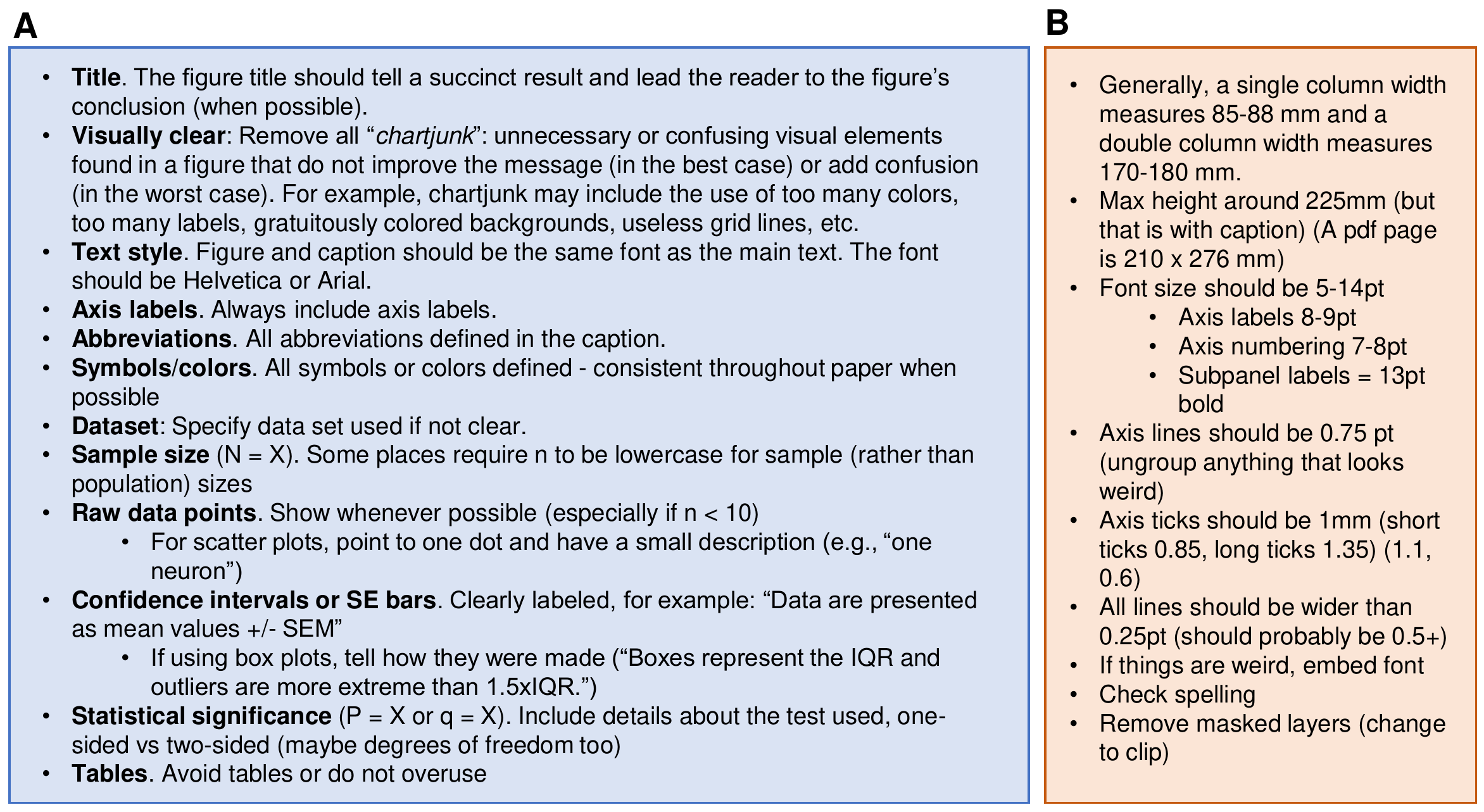}{1}
    {Making good figures requires attention to detail.} 
    {{
        {General principals of making and labelling a good figure. These should help facilitate understanding and reproducibility. }, 
        {Some more detailed technical specifics about making figures and their sizes. Inkscape is a great free vector figure editor and alternative to Adobe Illustrator.\captionTips} 
    }}

\section{Scientific Background}
\textbf{How do different choices of low complexity models help us understand and explain the generalization of a neural network? }
\textbf{Do neural networks, under certain hyperparameters, generalize through a self averaging effect as in boosting?}
\subsection{Motivation} In recent years, neural networks have made giant leaps in a wide variety of domains. Neural networks are often referred to as “black box” algorithms due to how little we can explain its empirical success. A lack of understanding raises concerns including but not limited to explainability, interpretability, adversarial robustness, and bias.

Our foundational research seeks to understand how and why neural networks generalize by using simpler, less complex models. Aristotle described one of the earliest forms of what we now call Occam’s Razor in his Posterior Analytics by saying “We may assume the superiority … [all else being equal] of the demonstration which derives from fewer postulates or hypotheses.” (Aristotle, Posterior Analytics, p. 150). From this theoretical framework, neural networks with lower complexity are preferred. In a practical setting, simple explanations for what a neural network learned and how it did so can bolster confidence for deployment as well as bring to light pitfalls.

\subsection{What’s Known} Developing a framework for understanding when and why neural networks work has been a longstanding problem. Early studies by \cite{kernel} found that NNs learn increasingly good representations in deeper layers, where good is defined as how much predictive signal we can extract (using PCA) from the associated Kernel matrix using a linear classifier. This understanding has proved valuable for tasks such as transfer learning. On the other hand, they also found that Neural networks trained with gradient descent struggle generalizing on few-shot “reasoning” tasks \cite{schmidhuber}. An example of such a task is predicting the number of on-bits in a 100-dimensional binary vector from just 3 examples. ML research on generalization today generally adopts gradient descent, instead studying the statistical assumptions/interpretations behind techniques of constraining the neural network. Common notions of “complexity” are in terms of the number and magnitude of parameters. 

One measure for complexity is just to measure the euclidean distance between a vector of all the weights of a model and a vector of all ones. This relates to the more classical measures like Kolmogorov complexity, where a neural network with most weights equal to zero have lower complexity due to fewer bits needed to store them. A paper by \cite{persistence} investigated an improvement on this by defining a topologically derived neural persistence measure based on the idea that some neurons contribute more to the final prediction. This work showed that finding the right measures of complexity can explain best practices in machine learning. In their experiments, dropout, a widely used technique for regularization, increased the neural persistence by a statistically significant amount. 

Another measure of complexity that is not specific to neural networks is the VC dimension, a measure of the capacity of a set of functions that can be learned. This general measure is well-studied for many function classes that we consider as candidates for “simple models”, from boosting classifiers to neural networks. For boosting, we know $d_k < 2kd \log(ke)$ where $d$ is the VC dimension of the base hypothesis class, and $d_k$ is that of the boosted hypothesis class after $k$ rounds of boosting. Moreover, the authors of Adaboost \cite{adaboost} gave various bounds on the generalization error in terms of the training error and $d_k$. Bounds on the VC dimension of MLPs have also been given for various choices of activations by \cite{shalev}. For example, with sigmoid activation, the VC dimension of a MLP with neurons V and connections E is $\Theta (|E|^2)$ and $O(|E|^2|V|^2)$. Thus, the VC dimension has the advantage of being general to any function class, and will be part of one of our conjectures relating boosted classifiers and neural networks.

Adaptive boosting was introduced by \cite{adaboost} as an improved boosting algorithm which adjusts adaptively to the errors of weak hypotheses found in previous iterations. Boosting has been observed to rarely overfit in practice, and many recent explanations have been proposed to explain why. These include margin-based explanations by \cite{margin} in terms of “mirror descent applied to the problem of maximizing the smallest margin in the training set under suitable separability conditions,” and they include optimization explanations in terms of the minimization of a convex loss function or “a procedure to search through the space of convex combinations of weak learners or base classifiers.” Notably, \cite{wyner} and \cite{mease} suggested that Adaboost exhibits the self-averaging mechanism of many interpolating classifiers and thus generalizes for the same reason random forests do. 

	More recently, it was shown by \cite{nakkiran} that NNs trained with stochastic gradient descent (SGD) share high performance correlation with linear classifiers in the initial phase of training and retain it afterwards. Contemporary work by \cite{tengyu} showed that the degree to which this is true depends on the learning rate. With a small learning rate the model learns a complex, non-linear decision boundary because it will memorize complex patterns wherever possible. They demonstrated this qualitatively by adding a memorizable patch to CIFAR-10 images, and showing the small learning rate model maximizes patch accuracy from the start. Their claims rely on the fact neural networks have a non-convex loss landscape, and may not hold outside the setting of SGD and a sufficiently large learning rate.

\subsection{What’s Still Missing} As described in \cite{nakkiran}’s paper, there are gaping holes in their central conjecture. Namely, there exists a “sequence” of “increasingly complex” functions $g_i$ and timesteps $T_i$ for which up until $T_i$, the neural network F learns nothing beyond $g_i$ and after $T_i$, retains it. Neither the function sequence or notion of complexity is defined. The combination of boosting and VC dimension have not been investigated as candidates to define this conjecture. 

	More broadly, the machine learning community can’t always predict in understandable terms when a model will or will not generalize. Measuring the performance correlation of two models is a new line of inquiry in this space, and has yet to be explored with other architectures than linear models. 

	To address the stochasticity of training, we provide a theoretically motivated training setup which corroborates our claims in two ways: 1) a high performance correlation between neural networks and boosted classifiers and 2) a similar distribution of errors over examples. We also formalize a general connection between the self-averaging property in dropout and boosted classifiers, and include in our analysis the degrees of sensitivity of this conjecture to various choices of hyperparameters.

    \clearpage 
    \section{Suggested Research}
\newtheorem{theorem}{Theorem}
\newtheorem{corollary}{Corollary}[theorem]
\newtheorem{conjecture}{Conjecture}

\subsection{Meta-goals} Explaining the generalization of all neural networks by a simple model class may be intractable (otherwise, there’d be no need for neural networks). Instead, we think different choices of “low complexity models” can better explain different constrained subclasses of neural networks. We want to identify one such model class and the corresponding subclass of neural networks.
More formally, we want to explain what a NN is learning at each phase of training corresponding to training iterations $T_i, T_{i+1}$ for $i >= 0$. Denote $F_i$ as the neural network at $T_i$. We want to find functions $G_i : i >= 0$ of increasing “complexity” such that $I(F_i ; Y | G_{i+1}) \simeq 0$, and $I(G_i ; Y | F_{i+1}) \simeq 0$ for all $i$. To fill in the missing parts, we want to materialize $G_i : i >= 0$, and try suitable definitions of “complexity”. A stretch goal is to also characterize the timesteps $T_i$, so we know when we can expect neural networks to generalize.
A third meta-goal which we describe in detail later is to “explain” what $F$ is learning. We can do so if we can see if $F$ learns increasingly “difficult” examples in the same order as $G_i$. 

\subsection{Specific Approaches}
\textbf{Boosting} Under VC dimension bounds given on Adaboost by \cite{margin} and \cite{adaboost}, we can formalize the complexity of the weak learners by varying the number of boosting iterations. Towards achieving the meta-goal about a series of increasingly complex learners, we can simultaneously train a neural network and Adaboost, checkpointing (saving) the neural network periodically. Here, the series of complex learners are boosted algorithms with more learners. Similarly, we checkpoint the final boosted classifier after each boosting round. Afterwards, we can plot the relevant quantities $I(F ; Y | G)$ and $I(G ; Y | F)$ for all pairwise combinations $(F, G)$ over the checkpoints. We expect to see a similar phase separation as in \cite{nakkiran}’s paper.

\textbf{Learning Order} It would go a long way to “explaining” what $F$ is learning if we can see if it’s learning increasingly “difficult” examples in the same order as $G_i$. To do so, we can define the difficulty of an example as the difficulty of the class it resides in, and its euclidean distance of its embedding from the mean of embeddings in that class. Then, we can measure progressive error on examples of various difficulty levels, and compare that between the neural network and boosted classifiers. In particular, we can take the embedding of examples, obtained using the pre-final layer. This can lead to a qualitative measure to explain similarity in how the model learns. The difficulties of the classes for common benchmark datasets have been explored in other works. As shown by \cite{hanxu}, different classes in CIFAR-10 are more or less difficult. As shown by \cite{tailin}, “discontinuous” phase transitions are observed by sweeping the Beta parameter of a binary classification loss function, which controls between compression (simplicity) and accuracy. These phase transitions correspond to new classes of examples the model learns. Plotting the accuracy against Beta shows many discontinuities, at which the model’s accuracy suddenly “jumps” upon learning either to recognize a new class, or to discriminate between samples of two visually similar classes. Thus, we may use the order of classes as a proxy for difficulty. Afterwards, we can also see if the times at which these phase transitions occur for a training neural network map onto the times at which the neural network correlates with increasingly complex functions $G_i$ , as will be defined in Theorem 1.

\subsection{Specific Questions} 

We plan to directly extend \cite{nakkiran}’s conjecture. We will replace g with a recursive definition of a class of functions, and fill in the notion of complexity as the VC dimension while keeping the same mutual performance correlation. 

\begin{theorem}

Let $G_i  = \sum_{j=1}^i h_j$, where all $h_j \in H$ (hypothesis class), and each $h_j$ is the (empirical risk minimizer) ERM on $S_i \sim P_i, D$. Let $F$ be $\{f: f(x) = \sum_{j=1}^k v_j f_j(x), \text{ and } f_j(x) = ReLU(\sum_{i=1}^d w_{i,j} x_i)\}$ denoting the set of one-hidden-layer neural networks with k hidden units and ReLU activations. Denote $F_i$ as the neural network at $T_i$. Then

\begin{equation}
    I(F_i ; Y | G_(i+1)) \simeq 0
\end{equation}
and
\begin{equation}
    I(G_i ; Y | F_(i+1)) \simeq 0
\end{equation}
for all $i$ from $1$ to $J$, and is statistically significant compared to increasingly better random classifiers $[R_i : 1 <= i <= J]$, that satisfy $I(R_i; Y) = I(F_i; Y)$.

\end{theorem}

Note that $S_i \sim P_i, D$ corresponds to the sample reweighting scheme in Adaboost, formalized drawing training samples from a different (but not independent) probability distribution each round. This is a reformulation of \cite{nakkiran}’s conjecture that is more direct: $F_i$ will learn $Y$ in the same way as $G_i$  then retain it afterwards as it learns the more complex $G_{i+1}$. That is equivalent to stating $F_i$ will never be more useful than $G_{i+1}$ in explaining $Y$ for all $i$. Vice-versa, $G_i$  will never be more useful than $F_{i+1}$ in explaining $Y$, stating $F$ has learned and fully retained $G_i$ .

To establish this relation, we have to draw a connection between the iterations of boosting in adaptive boosting, and sub-networks from the neural network. We do establish this relation in theorem 2.

\begin{theorem}
Let $G_i = \sum_{j}^i h_j$, where all $h_j \in H $  (hypothesis class), and each $h_j$ minimizes risk on $S_i \sim P_i, D$. Let $f(x) = \sum_{j}^k v_j f_j(x)$, where $f_j(x) = \text{ReLU}(\sum_{i=1}^d w_{i,j} x_i)$ denote a one-hidden-layer neural network $f$ with $k$ hidden units and ReLU activations. Indexed $f, h$ are random variables. SGD minibatch gradient descent will progressively increase and decrease 

\[\text{corr}(f_{i1}, h_{j1})  \text{and corr}(f_{i1}, f_{i2})\]
under some matching $\{i1, i2, …\} \text{ to } \{j1, j2, …\}$, respectively. Next, we
define $J \subset \{1, ..., k\}$, and $f^J(x) = \sum_{j \in J} v_j f_j(x)$ as a “sub-network” of $f$. SGD gradient descent will progressively decrease \[corr(f^J_{i1}(x), f^J_{i2}(x))\].
\end{theorem}

Note that this theorem, if true, will mean each $f_j$ comes up with an independent “hypothesis” which combines via a self-averaging mechanism to form the final hypothesis. Implicitly, this means each neuron $f_j$ will have to focus on different feature(s). Like in random forests and Adaboost, this self-averaging mechanism of relatively independent base hypotheses can explain why neural networks, under the common practice of dropout, can generalize well.

As pointed out by \cite{tengyu}’s paper, the dataset noise/signal matters. A specific question we also want to ask is if the mutual information that is shared comes from a specific portion of the data. To do so, we can experimentally separate the data in, say, CIFAR-100, into easy and difficult tasks as described in the learning order section. Then, we can plot the mutual information on each of these subsets of data and find ones that correlate. 

To make these sub-networks useful for practitioners trying to explain a NN with boosted classifiers, we make the following conjecture. 

\begin{conjecture}
Let $F(x)$ denote a one-hidden-layer neural network $f$ with $k$ hidden units and ReLU activations. There exists some mapping $\pi : \mathds{N} \to \mathds{N}$ such that we can find $G$ of VC dimension $\pi (\text{VCDim}(F))$ for which theorem 1 and 2 hold in a variety of settings.
\end{conjecture}
In boosting, one needs to choose both the base hypothesis class and the number of boosting rounds. With this conjecture, we envision general practices to choose the base hypothesis class. This will substantially accelerate the practitioner's' search for finding a boosted classifier which can pair with the black box neural network using the relationship between their VC dimensions.

\subsection{Tools and Techniques you intend to employ and develop} For the experimental part of our research, we will apply computer vision architectures such as CNN’s to CIFAR-100 data, as well as create our own synthetic data as in \cite{nakkiran}’s paper. The CNN will serve as our complex deep learning model which we wish to explain and study.

We will use pre-trained CNN’s like VGG-16 and feature extractors like ScatterNet to both extract features for our boosted classifiers and measure the “difficulty” of examples. Measuring the difficulty of an example will involve finding a combination of how much it deviates from others in its class and how close it is to other classes.

We intend to adopt proof techniques similar to \cite{nakkiran}, \cite{tengyu}, and others in the field. There, they precisely characterize the data distribution, architecture, and training procedure. Common defaults for the setup involve a data distribution with separable noise and signal components. A MLP with a single hidden layer with hidden layer ReLU activation and sigmoid or tanh final activation is commonly used for the architecture. Training procedure often involves feature standardization and uses the SGD optimizer. For proofs, hinge loss is often used.

We plan to use Adaboost due to its state-of-the-art results amongst boosting algorithms and ease of implementation. We will use the same measure of mutual information and plot this over the training time of our CNN expert model. 

\subsubsection{Experiment 1}
We will train and checkpoint a typical CNN to greater than 90 percent performance on CIFAR-10 first 5 versus last 5 binary classification, as well as on synthetic high dimensional sinusoidal data where we can control the ratio of signal to noise. Next, we will train and checkpoint adaboost at each round of boosting, where we experiment with standard weak learners and extracted features from VGG-16, as well as shallow CNN's on the raw images as in \cite{TAHERKHANI2020351}. We can qualitatively examine the plots of mutual information as in figure 1 to see observed phase separation.
\begin{figure}
    \centering
    \includegraphics[scale=0.8]{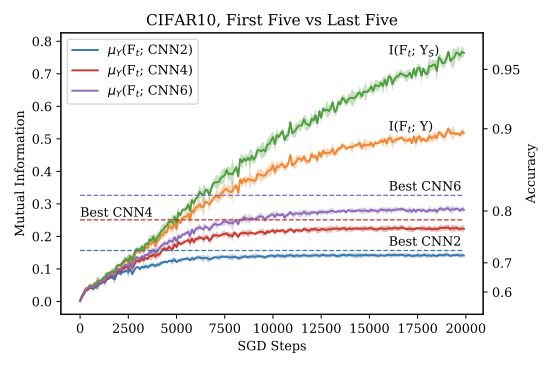}
    \caption{The plots generated by \cite{nakkiran} to qualitatively examine phase separation} 
    \label{fig:phases}
\end{figure}

Our quantitative analysis will involve checking that both quantities
\begin{equation}
    I(G_i; Y | F_{i+1}) \text{ and }.  
    I(F_i;Y|G_{i+1})
\end{equation}
are close to zero. We will optimize the number of training epochs at which each $F_i$ is defined for these two quantities to be small. Next, will measure the statistical significance of the phase separation by averaging over the quantities for many random classifiers of the same accuracy as $G_i$. To construct a random classifier with accuracy $a$ we output the correct label with probability $a$ and the incorrect label otherwise.

\subsubsection{Experiment 2}
Let errors($h_j$) and errors($f_j$) denote the emprical distribution of errors (e.g. $|f_j(x) - y| \quad \forall (x, y) \in X \times Y$) over the dataset using the sub-classifier $f_j$ directly to predict $X$. Aside from showing the correlation between f and h to decrease per our theoretical analysis, we’re also interested in looking at examples they make the same errors on. Specifically, we expect the distribution of errors to be similar between the NN sub-classifier $f_j$ and the weak classifier $h_j$ it matches with, and the distribution of errors to diverge between two different sub-classifiers $f_{j1}$ and $f_{j2}$, whose votes become more independent. Specifically, we can show that after phase $i$, there is a sudden decrease in $D_{KL}$(errors($f_j$), errors($h_j$)) and a sudden increase in $D_{KL}$(errors($f_{j1}$), errors($f_{j2}$)).

Lastly, we make sure the final test error of a NN constrained this way still compares favorably to a normally trained NN.

\subsection{Initial Ideas}

\subsubsection{Theorem 1} We think we can inductively establish the result, given two assumptions. First, $H^T$ is expressive, i.e. the data distribution and $H$ are suitably chosen so $\argmax_{f \in H^T} l(f(X), Y) \approx \argmax_{f \in F} l(f(X), Y)$ where $l$ denotes the likelihood objective. Second, the batch used to train $F_i$ in phase $i$ is drawn from the distribution $S_i$, which was used to fit weak classifier $i$. This will consist of the base case, where we show the neural network learns a low-complexity weak learner, drawn from some hypothesis class $H$. \cite{nakkiran} among others already show this for $H$ being the full linear classifier. Our base case will be no more difficult than that, since we sample the batch from the same observation weight distribution chosen by Adaboost. A proof sketch will be to show that $I(F_i ; Y | G_{i+1}) \simeq 0 \cap F_{i+1} \leftarrow \text{train}(F_i, S_i) \Rightarrow I(F_{i+1} ; Y | G_{i+2}) \simeq 0$ and similarly $I(G_i ; Y | F_{i+1}) \simeq 0 \cap F_{i+1} \leftarrow \text{train}(F_i, S_i) \Rightarrow I(G_{i+1} ; Y | F_{i+2}) \simeq 0$. Intuitively, $F_i$ and $h_i$ will minimize risk on the same examples. In Theorem 2, we will further extend this correspondence by finding a matching between $F_i$'s "sub-classifiers" and the weak classifiers $h_j$'s.

\subsubsection{Theorem 2} 

In the neuron view of the sub-classifier, we think we can create a correspondence between each neuron $f_j$ to a weak classifier $h_j$ by adopting the arguments in \cite{niladri}, which shows a single step of gradient descent can create "neuron alignment": a substantial number of neurons will correlate to some cluster mean in their chosen xor-like data distribution, hence activating on examples chosen from that cluster. They also show "almost-orthogonality": an aligned neuron won't activate on other cluster means (i.e. the complement). An idealized interpretation of their result in our case is that $x\in S_j \Leftrightarrow f_j(x) \approx 1$ and $x\not\in S_j \Leftrightarrow f_j(x) \approx 0$. 

In the subnetwork view of the sub-classifier, we think we can adopt a similar approach, but analysis may call for some additional assumptions. Instead of choosing an ensemble of subnetworks at random (similar to what dropout will do), we impose we get to pick them such that the number of shared neurons between any pair of them is small. By limiting to a small collection of subnetworks, and training each on different sets of observation weights informed by Adaboost, we can create a 1:1 correspondence between a subnetwork and a weak classifier. Extending the assumption from Theorem 1, each sub-classifier $f_j$ or $f^{J_j}$ will be trained batches sampled from $S_j$, so it learns using the same observations as a corresponding weak classifier, specializing to the same region.

Then, we will combine the aligned neurons (or ensemble of sub-networks) into a single sub-network. \cite{niladri} shows this creates a large margin sub-network (the substantial number of aligned neurons comprising its hidden layer) which results in a large margin classifier overall. This large margin sub-network’s decision implicitly averages over these sub-classifiers’ votes, hence it has the same self-averaging explanation as a boosted classifier!

\subsubsection{Conjecture 1}

To really make the connection hold in a wide variety of settings, we want to find $G$ given $F$ for which both hypotheses hold. Our theoretical analysis has assumed so far Adaboost yields a near-optimal classifier $G$ for which we then constrain the training of $F$ in order for it to be explained by $G$ both from correlation and causation. However, to truly deploy this into the real world, we would need to construct both in parallel, and this conjecture formulates our intuition that there’s a mapping between their complexities so that given one, we can narrow down the search for the other.

\subsubsection{Summary}

Whereas there is little work explaining both the complexity and performance gains of neural networks with every layer added, there is plenty of literature on VC dimension bounds and generalization error guarantees for boosting from PAC learning like \cite{adaboost}, \cite{margin} and much more. We want to see if the claims of \cite{nakkiran} still hold when a linear classifier is swapped with boosted classifiers and with VC dimension as the notion of complexity. 

We propose to first establish an explanation using correlation (using a mutual information based correlation measure through phases of training), then an explanation using causation by a shared self-averaging mechanism over sub-classifiers. We give experimental setups that can put these falsifiable hypotheses to the test. If true, that can go a long way in uniting two traditionally separate fields of research – boosting and deep learning – and helping to ground the success of neural networks in more explored territory. As a future direction, we envision when the situation calls for it, we can readily swap a neural network black box decision with an alternative that takes a small step down in accuracy but a giant leap forward in being explainable in the features and interpretable in their decisions.

For additional visualizations and toy cases that complement this proposal, one is welcome to reference our \href{https://docs.google.com/presentation/d/1eMMTieuvW1jJMDWejJZlGbEGDhYqXS9VqF9Co4aR7bs/edit?usp=sharing}{recent presentation}.

    \clearpage 
    \printbibliography 
    \clearpage


\end{document}